\documentclass[11pt,a4paper]{article}
\PassOptionsToPackage{breaklinks}{hyperref}
\usepackage[hyperref]{emnlp2018}
\usepackage{times}
\usepackage{latexsym}
\usepackage{xcolor}

\usepackage[utf8]{inputenc}
\usepackage[T1]{fontenc}
\usepackage{graphicx}
\usepackage{breakurl}
\usepackage{multirow}

\newenvironment{citemize}{\begin{list}{$\bullet$}{\topsep=.1\smallskipamount\itemsep=0pt\parsep=1pt\labelwidth=.5em}}{\end{list}}
\aclfinalcopy 

\title{Grammatical Error Correction in Low-Resource Scenarios}

\author{Jakub N\'{a}plava \and Milan Straka\\
  Charles University, \\
  Faculty of Mathematics and Physics, \\
  Institute of Formal and Applied Linguistics \\
  \texttt{\{naplava,straka\}@ufal.mff.cuni.cz}}

\date{}

\begin{document}
\maketitle
\begin{abstract}
  Grammatical error correction in English is a~long studied problem with many existing systems and datasets. However, there has been only a limited research on error correction of other languages. In this paper, we present a~new dataset AKCES-GEC on grammatical error correction for Czech. We then make experiments on Czech, German and Russian and show that when utilizing synthetic parallel corpus, Transformer neural machine translation model can reach new state-of-the-art results on these datasets. AKCES-GEC is published under CC BY-NC-SA 4.0 license at \url{http://hdl.handle.net/11234/1-3057}, and the source code of the GEC model is available at \url{https://github.com/ufal/low-resource-gec-wnut2019}.
\end{abstract}

\section{Introduction}

A great progress has been recently achieved in grammatical error correction (GEC) in English. The performance of systems has since CoNLL 2014 shared task~\cite{ng2014conll} increased by more than 60\% on its test set~\cite{bryant2019bea} and also a variety of new datasets appeared. Both rule-based models, single error-type classifiers and their combinations were due to larger amount of data surpassed by statistical and later by neural machine translation systems. These address GEC as a translation problem from a language of ungrammatical sentences to a grammatically correct ones. 

Machine translation systems require large amount of data for training. To cope with this issue, different approaches were explored, from acquiring additional corpora (e.g. from Wikipedia edits) to building a synthetic corpus from clean monolingual data. This was apparent on recent Building Educational Applications (BEA) 2019 Shared Task on GEC~\cite{bryant2019bea} when top scoring teams extensively utilized synthetic corpora.

The majority of research has been done in English. Unfortunately, there is a limited progress on other languages. Namely, \newcite{boyd2018using} created a dataset and presented a GEC system for German, \newcite{rozovskaya2019grammar} for Russian, \newcite{naplava2017natural} for Czech and efforts to create annotated learner corpora were also done for Chinese~\cite{yu2014overview}, Japanese~\cite{mizumoto2011mining} and Arabic~\cite{zaghouani2015large}. 

Our contributions are as follows:
\begin{citemize}
  \item We introduce a new Czech dataset for GEC. In comparison to dataset of \newcite{CzeSL_GEC_1} it contains separated edits together with their type annotations in M2 format~\cite{dahlmeier2012better} and also has two times more sentences.
  \item We extend the GEC model of \newcite{naplava2019cuni} by utilizing synthetic training data, and evaluate it on Czech, German and Russian, achieving state-of-the-art results.
\end{citemize}


\section{Related Work}

There are several main approaches to GEC in \emph{low-resource} scenarios. The first one is based on a noisy channel model and consists of three components: a candidate model to propose (word) alternatives, an error model to score their likelihood and a language model to score both candidate (word) probability and probability of a whole new sentence. \newcite{richter2012korektor} consider for a given word all its small modifications (up to character edit distance 2) present in a morphological dictionary. The error model weights every character edit by a trained weight, and three language models (for word forms, lemmas and POS tags) are used to choose the most probable sequence of corrections. A candidate model of \newcite{bryant2018language} contains for each word spell-checker proposals, its morphological variants (if found in Automatically Generated Inflection Database) and, if the word is either preposition or article, also a set of predefined alternatives. They assign uniform probability to all changes, but use strong language model to re-rank all candidate sentences. \newcite{lacroix2019noisy} also consider single word edits extracted from Wikipedia revisions.

Other popular approach is to extract parallel sentences from Wikipedia revision histories. A great advantage of such an approach is that the resulting corpus is, especially for English, of great size. However, as Wikipedia edits are not human curated specifically for GEC edits, the corpus is extremely noisy. \newcite{grundkiewicz2014wiked} filter this corpus by a set of regular expressions derived from NUCLE training data and report a performance boost in statistical machine translation approach. \newcite{grundkiewicz2019neural} filter Wikipedia edits by a simple language model trained on BEA 2019 development corpus. \newcite{lichtarge2019corpora}, on the other hand, reports that even without any sophisticated filtering, Transformer~\cite{vaswani2017attention} can reach surprisingly good results when used iteratively.

The third approach is to create synthetic corpus from a clean monolingual corpus and use it as additional data for training. Noise is typically introduced either by rule-based substitutions or by using a subset of the following operations: token replacement, token deletion, token insertion, multi-token swap and spelling noise introduction. \newcite{yuan2013constrained} extract edits from NUCLE and apply them on a clean text. \newcite{choe2019neural} apply edits from W\&I+Locness training set and also define manual noising scenarios for preposition, nouns and verbs. \newcite{zhao2019improving} use an unsupervised approach to synthesize noisy sentences and allow deleting a word, inserting a random word, replacing a word with random word and also shuffling (rather locally). \newcite{grundkiewicz2019neural} improve this approach and replace a token with one of its spell-checker suggestions. They also introduce additional spelling noise.
 

\section{Data}

In this Section, we present existing corpora for GEC, together with newly released corpus for Czech.

\subsection{AKCES-GEC}

\begin{figure*}[ht!]
    \centering
    \includegraphics[width=\hsize]{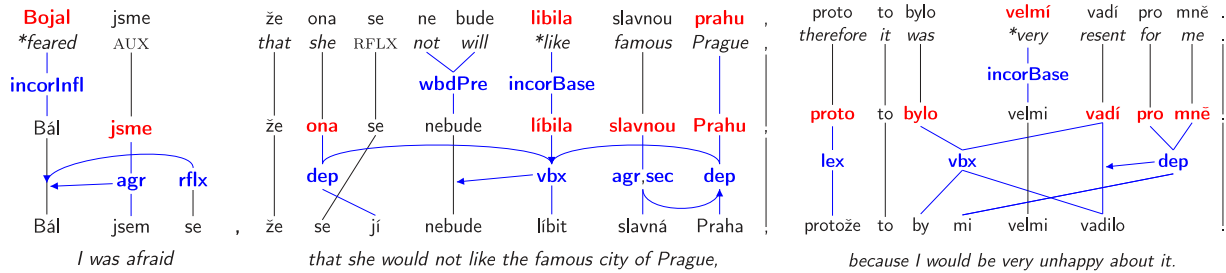}
    \caption{Example of two-level annotation of a sentence in CzeSL corpus, reproduced from \cite{Rosen:2016a}.}
    \label{fig:czesl_annot}
\end{figure*}

\begin{figure*}[ht!]
    \centering
    \includegraphics[width=.8\hsize]{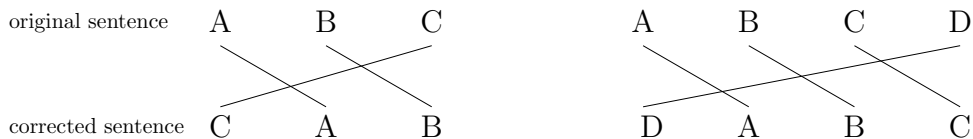}
    \caption{Word swap over one or two correct words (on the left) is considered
    a single edit (A B C$\rightarrow$ C A B). Word swap over more than
    two correct words (on the right) is represented as two edits of deleting D and inserting D.}
    \label{fig:merging_swaps}
\end{figure*}

\begin{table*}[ht!]
    \centering
    \small
    \setlength{\tabcolsep}{3.2pt}
    \begin{tabular}{l|l|l|r}
    Error type & Description & Example & Occ \\\hline
\emph{incorInfl} & incorrect inflection & [\textbf{pracovají} $\rightarrow$ \emph{pracují}] v továrně & 8\,986\\ 
\emph{incorBase} & incorrect word base & musíš to [\textbf{posvětlit} $\rightarrow$ \emph{posvětit}] & 20\,334 \\ 
\emph{fwFab} & non-emendable, „fabricated“ word & pokud nechceš slyšet [\textbf{smášky}] & 78 \\
\emph{fwNC} & foreign word & váza je na [\textbf{Tisch} $\rightarrow$ \emph{stole}] & 166 \\
\emph{flex} & supplementary flag used with fwFab and & jdu do [\textbf{shopa} $\rightarrow$ \emph{obchodu}] & 34 \\
      & fwNC marking the presence of inflection  & \\
\emph{wbdPre} & prefix separated by a space or preposition w/o space & musím to [\textbf{při pravit} $\rightarrow$ \emph{připravit}] & 817 \\
\emph{wbdComp} & wrongly separated compound & [\textbf{český anglický} $\rightarrow$ \emph{česko-anglický}] slovník & 92 \\
\emph{wbdOther} & other word boundary error  & [\textbf{mocdobře} $\rightarrow$ \emph{moc dobře}]; [\textbf{atak} $\rightarrow$ \emph{a tak}] & 1326 \\
\emph{stylColl} & colloquial form & [\textbf{dobrej} $\rightarrow$ \emph{dobrý}] film & 3\,533 \\
\emph{stylOther} & bookish, dialectal, slang, hyper-correct form & holka s [\textbf{hnědými očimi} $\rightarrow$ \emph{hnědýma očima}] & 156 \\
\hline

\emph{agr} & violated agreement rules & to jsou [\textbf{hezké} $\rightarrow$ \emph{hezcí}] chlapci; Jana [\textbf{čtu} $\rightarrow$ \emph{čte}] & 5\,162 \\
\emph{dep} & error in valency & bojí se [\textbf{pes} $\rightarrow$ \emph{psa}]; otázka [\textbf{čas} $\rightarrow$ \emph{času}] & 6\,733 \\ 
\emph{ref} & error in pronominal reference  & dal jsem to jemu i [\textbf{jejího} $\rightarrow$ \emph{jeho}] bratrovi & 344 \\
\emph{vbx} & error in analytical verb form or compound predicate & musíš [\textbf{přijdeš} $\rightarrow$ \emph{přijít}]; kluci [\textbf{jsou}] běhali & 864 \\ 
\emph{rflx} & error in reflexive expression & dívá [$\emptyset \rightarrow$ \emph{se}] na televizi; Pavel [\textbf{si} $\rightarrow$ \emph{se}] raduje & 915 \\
\emph{neg} & error in negation & [\textbf{půjdu ne} $\rightarrow$ \emph{nepůjdu}] do školy & 111\\
\emph{lex} & error in lexicon or phraseology & dopadlo to [\textbf{přírodně} $\rightarrow$ \emph{přirozeně}] & 3\,967 \\
\emph{use} & error in the use of a grammar category & pošta je [\textbf{nejvíc blízko} $\rightarrow$ \emph{nejblíže}]  & 1\,458 \\ 
\emph{sec} & secondary error (supplementary flag) & stará se o [\textbf{našich holčičkách} $\rightarrow$ \emph{naše holčičky}] & 866 \\
\emph{stylColl} & colloquial expression & viděli jsme [\textbf{hezký} $\rightarrow$ \emph{hezké}] holky & 3\,533 \\
\emph{stylOther} & bookish, dialectal, slang, hyper-correct expression & rozbil se mi [\textbf{hadr}] & 156 \\
\emph{stylMark} & redundant discourse marker & [\textbf{no}]; [\textbf{teda}]; [\textbf{jo}] & 15 \\
\emph{disr} & disrupted construction & známe [\textbf{hodné spoustu} $\rightarrow$ \emph{spoustu hodných}] lidí & 64 \\
\hline
\emph{problem} & supplementary label for problematic cases &  & 175 \\
\hline
\emph{unspec} & unspecified error type &  & 69\,123 \\
    \end{tabular}
    \caption{Error types used in CzeSL corpus taken from \cite{CzeSL:TSD2012}, including number of occurrences in the dataset being released. Tier 1 errors are in the upper part of the table, Tier 2 errors are in the lower part.
    The \textit{stylColl} and \textit{stylOther} are annotated on both Tiers, but we do not distinguish on which one
    in the AKCES-GEC.}
    \label{tab:czesl_errors}
\end{table*}

\begin{table*}[t]
    \centering
    \setlength{\tabcolsep}{2.1pt}
    \renewcommand{\arraystretch}{1.1}
    \begin{tabular}{ll||r|r|r|r||r|r|r|r||r|r|r|r}
    & & \multicolumn{4}{c||}{Train} & \multicolumn{4}{c||}{Dev} & \multicolumn{4}{c}{Test} \\\hline
    & & \multicolumn{1}{c|}{Doc} & \multicolumn{1}{c|}{Sent} & \multicolumn{1}{c|}{Word} & \multicolumn{1}{c||}{Error r.} & \multicolumn{1}{c|}{Doc} & \multicolumn{1}{c|}{Sent} & \multicolumn{1}{c|}{Word} & \multicolumn{1}{c||}{Error r.} & \multicolumn{1}{c|}{Doc} & \multicolumn{1}{c|}{Sent} & \multicolumn{1}{c|}{Word} & \multicolumn{1}{c}{Error r.} \\\hline\hline  
    \multirow{2}{*}{Foreign.} & Slavic     & \multirow{2}{*}{1\,816} & \multirow{2}{*}{27\,242} &     \multirow{2}{*}{289\,439} & \multirow{2}{*}{22.2~\%} &
      70 & 1\,161 & 14\,243 & 21.8~\% & 69 & 1\,255 & 14\,984 & 18.8~\% \\\cline{2-2}\cline{7-14}
                                & Other & & & & &
      45 &    804 &  8\,331 & 23.8~\% & 45 &    879 &  9\,624 & 20.5~\% \\\hline
    \multicolumn{2}{l||}{Romani}              & 1\,937 & 14\,968 & 157\,342 & 20.4~\% &
      80 &  520 &    5\,481 & 21.0~\% & 74 &    542 &  5\,831 & 17.8~\% \\\hline\hline
    \multicolumn{2}{l||}{Total}               & 3\,753 & 42\,210 & 446\,781 & 21.5~\% &
     195 & 2\,485 & 28\,055 & 22.2~\% & 188 & 2\,676 & 30\,439 & 19.1~\% \\
    \end{tabular}
    \caption{Statistics of the AKCES-GEC dataset -- number of documents, sentences, words and error rates.}
    \label{tab:akcesgec_sizes}
\end{table*}

The AKCES (Czech Language Acquisition Corpora; \citealp{sebesta2010}) is an umbrella project comprising of several acquisition resources -- CzeSL (learner corpus of Czech as a second language), ROMi (Romani ethnolect of Czech Romani children and teenagers) and SKRIPT and SCHOLA (written and spoken language collected from native Czech pupils, respectively).

We present the AKCES-GEC dataset, which is a~grammar error correction corpus for Czech generated from a subset of AKCES resources. Concretely, the AKCES-GEC dataset is based on CzeSL-man corpus \cite{Rosen:2016a} consisting of manually annotated transcripts of essays of non-native speakers of Czech. Apart from the released CzeSL-man, AKCES-GEC further utilizes additional unreleased parts of CzeSL-man and also essays of Romani pupils with Romani ethnolect of Czech as their first language.

The CzeSL-man annotation consists of three Tiers -- Tier 0 are transcribed inputs, followed by the level of orthographic and morphemic corrections, where only word forms incorrect in any context are considered (Tier 1). Finally, the rest of errors is annotated at Tier 2. Forms at different Tiers are manually aligned and can be assigned one or more error types \cite{CzeSL:TSD2012}. An example of the annotation is presented in Figure~\ref{fig:czesl_annot}, and the list of error types used in CzeSL-man annotation is listed in Table~\ref{tab:czesl_errors}.

We generated AKCES-GEC dataset using the three Tier annotation of the underlying corpus. We employed Tier 0 as source texts, Tier 2 as corrected texts, and created error edits according to the manual alignments, keeping error annotations where available.\footnote{The error annotations are unfortunately not available in the whole underlying corpus, and not all errors are annotated with at least one label.} Considering that the M2 format \cite{dahlmeier2012better} we wanted to use does not support non-local error edits and therefore cannot efficiently encode word transposition on long distances, we decided to consider word swaps over at most 2 correct words a single edit (with the constant 2 chosen according to the coverage of long-range transpositions in the data). For illustration, see Figure~\ref{fig:merging_swaps}.

The AKCES-GEC dataset consists of an explicit train/development/test split, with each set divided into foreigner and Romani students; for development and test sets, the foreigners are further split into Slavic and non-Slavic speakers. Furthermore, the development and test sets were annotated by two annotators, so we provide two references if the annotators utilized the same sentence segmentation and produced different annotations.

The detailed statistics of the dataset are presented in Table~\ref{tab:akcesgec_sizes}.
The AKCES-GEC dataset is released under the CC BY-NC-SA 4.0 license
at {\small\url{http://hdl.handle.net/11234/1-3057}}.

We note that there already exists a CzeSL-GEC dataset \cite{CzeSL_GEC_1}. However, it consists only of
a subset of data and does not contain error types nor M2 files with individual edits.

\subsection{English}

Probably the largest corpus for English GEC is the Lang-8 Corpus of Learner English \citep{mizumoto2011mining, tajiri2012tense}. It comes from an online language learning website, where users are able to post texts in language they are learning. These texts then appear to native speakers for correction. The corpus has over 100\,000 raw English entries comprising of more than 1M sentences. Due to the fact that texts are corrected by online users, this corpus is also quite noisy.

Other corpora are corrected by trained annotators making them much cleaner but also significantly smaller. NUCLE~\cite{dahlmeier2013building} has 57\,151 sentences originating from 1\,400 essays written by mainly Asian undergraduate students at the National University of Singapore. FCE~\cite{yannakoudakis2011new} is a subset of the Cambridge Learner
Corpus (CLC) and has 33\,236 sentences from 1\,244 written answers to FCE exam questions. Recent Write \& Improve (W\&I) and LOCNESS v2.1 \citep{bryant2019bea, granger1998} datasets were annotated for different English proficiency levels and a part of them also comes from texts written by native English speakers. Altogether, it has 43\,169 sentences.

To evaluate system performance, CoNLL-2014 test set is most commonly used. It comprises of 1\,312 sentences written by 25 South-East Asian undergraduates. The gold annotations are matched against system hypothesis using MaxMatch scorer outputting $F_{0.5}$ score. The other frequently used dataset is JFLEG~\cite{napoles-sakaguchi-tetreault:2017:EACLshort, heilman-EtAl:2014:P14-2}, which also tests systems for how fluent they sound by utilizing the GLEU metric~\cite{napoles2015ground}. Finally, recent W\&I and LOCNESS v2.1 test set allows to evaluate systems on different levels of proficiency and also against different error types (utilizing ERRANT scorer).

\subsection{German}

\newcite{boyd2018using} created GEC corpus for German from two German learner corpora: Falko and MERLIN~\cite{boyd2014merlin}. The resulting dataset comprises of 24\,077 sentences divided into training, development and test set in the ratio of 80:10:10. To evaluate system performance, MaxMatch scorer is used.

Apart from creating the dataset, \newcite{boyd2018using} also extended ERRANT for German. She defined 21 error types (15 based on POS tags) and extended spaCy\footnote{\scriptsize\url{https://spacy.io/}} pipeline to classify them.

\subsection{Russian}

\newcite{rozovskaya2019grammar} introduced RULEC-GEC dataset for Russian GEC. To create this dataset, a~subset of RULEC corpus with foreign and heritage speakers was corrected. The final dataset has 12\,480 sentences annotated with 23 error tags. The training, development and test sets contain 4\,980, 2\,500 and 5\,000 sentence pairs, respectively.

\subsection{Corpora Statistics}
 Table~\ref{tab:data_overview} indicates that there is a variety of English datasets for GEC. As \newcite{naplava2019cuni} show, training Transformer solely using these annotated data gives solid results. On the other hand, there is only limited number of data for Czech, German and Russian and also the existing systems perform substantially worse. This motivates our research in these low-resource languages.
 
 Table~\ref{tab:data_overview} also presents an average error rate of each corpus. It is computed using maximum alignment of original and annotated sentences as a ratio of non-matching alignment edges (insertion, deletion, and replacement). The highest error rate of 21.4 \% is on Czech dataset. This implies that circa every fifth word contains an error. German is also quite noisy with an error rate of 16.8 \%. The average error rate on English ranges from 6.6 \% to 14.1 \% and, finally, the Russian corpus contains the least errors with an average error rate of 6.4\%.

\begin{table}[t]
    \centering
    \setlength{\tabcolsep}{3.3pt}
    \begin{tabular}{l|l|r|r}
        \multicolumn{1}{c|}{Language} & \multicolumn{1}{c|}{Corpus} & \multicolumn{1}{c|}{Sentences} & \multicolumn{1}{c}{Err. r.} \\\hline
        \multirow{3}{*}{English} & Lang-8 & 1\,147\,451 &  14.1\% \\\cline{2-4} 
                                 & NUCLE & 57\,151 & 6.6\%\\\cline{2-4} 
                                 & FCE & 33\,236  & 11.5\% \\\cline{2-4} 
                                 & W\&I+LOCNESS & 43\,169 & 11.8\% \\\hline 
        Czech & AKCES-GEC & 42\,210 & 21.4\% \\\hline 
        German & Falko-MERLIN & 24\,077 & 16.8\% \\\hline 
        Russian & RULEC-GEC & 12\,480 & 6.4\% \\
    \end{tabular}
    \caption{Statistics of available corpora for Grammatical Error Correction.}
    \label{tab:data_overview}
\end{table}

\begin{table*}[t]
  \begin{center}
    \begin{tabular}{l|c|c|c|c|c||c|c|c|c|c}
     Language & \multicolumn{5}{c||}{Token-level operations} & \multicolumn{5}{c}{Character-level operations} \\\hline
     & sub & ins & del & swap & recase & sub & ins & del & recase & \raisebox{4pt}[13pt][8pt]{\vtop{\hbox{~~toggle}\hbox{diacritics}}} \\\hline
       English & 0.6~~ & 0.2~~ & 0.1~~ & 0.05 & 0.05 & 0.25 & 0.25 & 0.25 & 0.25 & 0~~~ \\\hline
      Czech & 0.7~~ & 0.1~~ & 0.05 & 0.1~~ & 0.05 & 0.2~~ & 0.2~~ & 0.2~~ & 0.2~~ & 0.2 \\\hline
      German & 0.64 & 0.2~~ & 0.1~~ & 0.01 & 0.05 & 0.25 & 0.25 & 0.25 & 0.25 & 0~~~  \\\hline
      Russian & 0.65 & 0.1~~ & 0.1~~ & 0.1~~ & 0.05 & 0.25 & 0.25 & 0.25 & 0.25 & 0~~~  \\\hline
    \end{tabular}
  \end{center}
  \caption{Language specific constants for token- and character-level noising operations.}
  \label{table:dataset_noisy_constants}
\end{table*}

\subsection{Tokenization}

The most popular metric for benchmarking systems are MaxMatch scorer~\cite{dahlmeier2012better} and ERRANT scorer~\cite{bryant2017automatic}. They both require data to be tokenized; therefore, most of the GEC datasets are tokenized.

To tokenize monolingual English and German data, we use spaCy v1.9.0 tokenizer utilizing \textit{en\_core\_web\_sm-1.2.0} and \textit{de} model. We use custom tokenizers for Czech\footnote{A slight modification of MorphoDiTa tokenizer.} and  Russian\footnote{\scriptsize\url{https://github.com/aatimofeev/spacy_russian_tokenizer}}.

\section{System Overview}

We use neural machine translation approach to GEC. Specifically, we utilize Transformer model~\cite{vaswani2017attention} to translate ungrammatical sentences to grammatically correct ones. We further follow \newcite{naplava2019cuni} and employ source and target word dropouts, edit-weighted MLE and checkpoint averaging. We do not use iterative decoding in this work, because it substantially slows down decoding. Our models are implemented in Tensor2Tensor framework version 1.12.0.\footnote{\scriptsize\url{https://github.com/tensorflow/tensor2tensor}}

\subsection{Pretraining on Synthetic Dataset}

Due to the limited number of annotated data in Czech, German and Russian we decided to create a corpus of synthetic parallel sentences. We were also motivated by the fact that such approach was shown to improve performance even in English with substantially more annotated training data. 

We follow \newcite{grundkiewicz2019neural}, who use an unsupervised approach to create noisy input sentences. Given a clean sentence, they sample a probability $p_{err\_word}$ from a normal distribution with a predefined mean and a standard deviation. After multiplying $p_{err\_word}$ by a number of words in the sentence, as many sentence words are selected for modification. For each chosen word, one of the following operations is performed with a predefined probability: substituting the word with one of its ASpell\footnote{\scriptsize\url{http://aspell.net/}} proposals, deleting it, swapping it with its right-adjacent neighbour or inserting a random word from dictionary after the current word. To make the system more robust to spelling errors, same operations are also used on individual characters with $p_{err\_char}$ sampled from a normal distribution with a different mean and standard deviation than $p_{err\_word}$ and (potentially) different probabilities of character operations.

When we inspected the results of a model trained on such dataset in Czech, we observed that the model often fails to correct casing errors and sometimes also errors in diacritics. Therefore, we extend word-level operations to also contain operation to change casing of a word. If a word is chosen for modification, it is with 50\% probability whole converted to lower-case, or several individual characters are chosen and their casing is inverted. To increase the number of errors in diacritics, we add a new character-level noising operation, which for a~selected character either generates one of its possible diacritized variants or removes diacritics. Note that this operation is performed only in Czech.

We generate synthetic corpus for each language from WMT News Crawl monolingual training data~\cite{ondrej2017findings}. We set $p_{err\_word}$ to 0.15, $p_{err\_char}$ to 0.02 and estimate error distributions of individual operations from development sets of each language. The constants used are presented in Table~\ref{table:dataset_noisy_constants}. We limited amount of synthetic sentences to 10M in each language.

\subsection{Finetuning}

A model is (pre-)trained on a synthetic dataset until convergence. Afterwards, we finetune the model on a mix of original language training data and synthetic data. When finetuning the model, we preserve all hyperparameters (e.g., learning rate and optimizer moments). In other words, the training continues and only the data are replaced.

When finetuning, we found that it is crucial to preserve some portion of synthetic data in the training corpus. Finetuning with original training data leads to fast overfitting with worse results on all of Czech, German and Russian. We also found out that it also slightly helps on English.

We ran a small grid-search to estimate the ratio of synthetic versus original sentences in the finetuning phase. Although the ratio of 1:2 (5M original oversampled training pairs and 10M synthetic pairs) still overfits, we found it to work best for English, Czech and German, and stop training when the performance on the development set starts deteriorating. For Russian, the ratio of 1:20 (0.5M oversampled training pairs and 10M synthetic pairs) works the best.

The original sentences for English finetuning are concatenated sentences from Lang-8 Corpus of Learner English, FCE, NUCLE and W\&I and LOCNESS. To better match domain of test data, we oversampled training set by adding W\&I training data 10 times, FCE data 5 times and NUCLE corpus 5 times to the training set. The original sentences in Czech, German and Russian are the training data of the corresponding languages.

\subsection{Implementation Details}

When running grid search for hyperparameter tuning, we use \textit{transformer\_base\_single\_gpu} configuration, which uses only 1 GPU to train \textit{Transformer Base} model. After we select all hyperparameter, we train \textit{Transformer Big} architecture on 4 GPUs. Hyperparameters described in following paragraphs belong to both architectures.

We use Adafactor optimizer~\cite{shazeer2018adafactor}, linearly increasing the learning rate from 0 to 0.011 over the first 8000 steps, then decrease it proportionally to the number of steps after that (using the \texttt{rsqrt\_decay} schedule). Note that this only applies to the pre-training phase.

All systems are trained on Nvidia P5000 GPUs. The vocabulary consists of approximately 32k most common word-pieces, the batch size is 2000 word-pieces per each GPU and all sentences with more than 150 word-pieces are discarded during training. Model checkpoints are saved every hour.

At evaluation time, we decode using a beam size of 4. Beam-search length-balance decoding hyperparameter alpha is set to 0.6. 

\section{Results}

\begin{table*}[t!]
  \begin{center}
    \begin{tabular}{l||c||c||c|c}
      \multirow{2}{*}{System} & \multirow{2}{*}{W\&I+L test} & \multirow{2}{*}{W\&I+L dev} & \multicolumn{2}{c}{CoNLL 14 test} \\\cline{4-5}
      & & & No W\&I+L & \kern-.4em With W\&I+L\kern-.4em \\\hline
      \multicolumn{5}{c}{including ensembles} \\\hline
      \newcite{lichtarge2019corpora} & -- & -- & 60.40 & -- \\\hline
      \newcite{zhao2019improving} & -- & -- & 61.15 & -- \\\hline
      \newcite{xu2019erroneous} & 67.21 & 55.37 & -- & 63.20 \\\hline
      \newcite{choe2019neural} & 69.06 & 52.79 & 57.50 & -- \\\hline
      \newcite{grundkiewicz2019neural} & \textbf{69.47} & 53.00  & \textbf{61.30} & \textbf{64.16} \\\hline
      \multicolumn{5}{c}{no ensembles} \\\hline
      \newcite{lichtarge2019corpora} & -- & -- & \textbf{56.80} & -- \\\hline
      \newcite{xu2019erroneous} & \textbf{63.94} & 52.29 & -- & 60.90 \\\hline
      \newcite{choe2019neural} & 63.05 & 47.75 & -- & -- \\\hline
      \newcite{grundkiewicz2019neural} & -- & 50.01 & -- & -- \\\hline\hline
      \multicolumn{5}{c}{no ensembles} \\\hline
      Our work -- synthetic pretrain & 51.16 & 32.76 & 41.85 & 44.12 \\\hline 
      Our work -- finetuned base single GPU & 67.18 & 52.80 & 59.87  & -- \\\hline 
      Our work -- finetuned & 69.00 & 53.30 & 60.76 & 63.40 \\\hline
       
    \end{tabular}
  \end{center}
  \caption{Comparison of systems on two English GEC datasets. CoNLL 2014 Test Set is divided into two system groups (columns): those who do not train on W\&I+L training data and those who do.}
  \label{table:english_results}
\end{table*}

\begin{table*}[t]
  \begin{center}
    \begin{tabular}{l||c||c||c}
      System & P & R  & $F_{0.5}$ \\\hline
      \newcite{boyd2018using} & 51.99 & 29.73 & 45.22 \\\hline\hline
      Our work -- synthetic pretrain  & 67.45 & 26.35 & 51.41 \\\hline 
      Our work -- finetuned base single GPU & 78.11 & 59.13 & 73.40 \\\hline 
      Our work -- finetuned & 78.21 & 59.94 & 73.71 \\\hline
       
    \end{tabular}
  \end{center}
  \caption{Results on on Falko-Merlin Test Set (German).}
  \label{table:german_results}
\end{table*}

\begin{table*}[t]
    \centering
    \begin{tabular}{l|l||c|c|c}
        \multicolumn{1}{c|}{System} & \multicolumn{1}{c|}{Test Subset} & \multicolumn{1}{c|}{P} & \multicolumn{1}{c|}{R} & \multicolumn{1}{c}{ $F_{0.5}$ } \\\hline
        \newcite{richter2012korektor} & All & 68.72 & 36.75 & 58.54 \\\hline\hline
        Our work -- synthetic pretrain & All & 80.32 & 39.55 & 66.59 \\\hline 
        Our work -- finetuned base single GPU & All & 84.21 & 66.67 & 80.00 \\\hline 
        \multirow{4}{*}{Our work -- finetuned} & Foreigners -- Slavic & 84.34 & 71.55 & 81.43 \\\cline{2-5}
        & Foreigners -- Other & 81.03 & 62.36 & 76.45 \\\cline{2-5}
        & Romani & 86.61 & 71.13 & 83.00 \\\cline{2-5}
        & All &  83.75 & 68.48 & 80.17 \\\hline
    \end{tabular}
    \caption{Results on on AKCES-GEC Test Set (Czech).}
    \label{table:czech_results}
\end{table*}

\begin{figure*}[t]
  \begin{center}
    \includegraphics[width=.7\hsize]{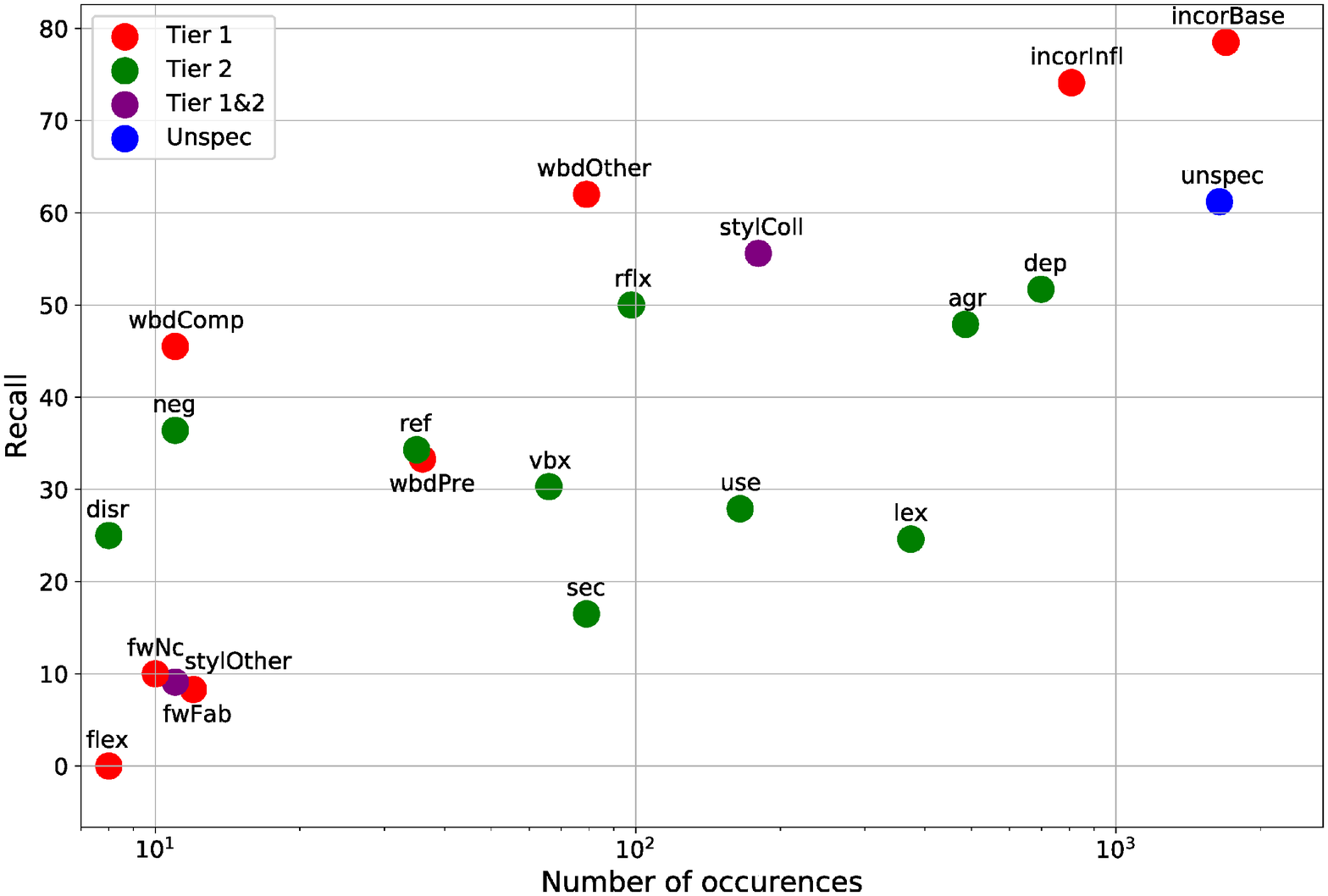}
  \end{center}
  \caption{Recall for each error type in the test set of AKCES-GEC, computed using the first annotator (ID 0).}
  \label{fig:czech_error_recalls}
\end{figure*}

We present results of our model when trained on English, Czech, German and Russian in this Section. As we are aware of only one system in German, Czech and Russian to compare with, we start with English model discussion. We show that our model is on par or even slightly better than current state-of-the-art systems in English when no ensembles are allowed. We then discuss our results on other languages, where our system exceeds all existing systems by a large margin.

In all experiments, we report results of three systems: \textit{synthetic pretrain}, which is based on Transformer Big and is trained using synthetic data only, and \textit{finetuned} and \textit{finetuned base single GPU}, which are based on Transformer Big and Base, respectively, and are both pretrained and finetuned. Note that even if the \textit{finetuned base} system has 3 times less parameters than \textit{finetuned}, its results on some languages are nearly identical.

We also tried training the system using annotated data only. With our model architecture, all but English experiments (which contain substantially more data) starts overfitting quickly, yielding poor performance. The overfitting problem could be possibly addressed as proposed by \newcite{sennrich2019revisiting}. Nevertheless, given that our best system on English is by circa 10 points in $F_{0.5}$ score better than the system trained solely on annotated data, we focused primarily on the synthetic data experiments. 

Apart from the W\&I+L development and test sets, which are evaluated using ERRANT scorer, we use MaxMatch scorer in all experiments.

\subsection{English}

We provide comparison between our model and existing systems on W\&I+L test and development sets and on CoNLL 14 test set in Table~\ref{table:english_results}. Even if the results on the W\&I+L development set are only partially indicative of system performance, we report them due to the W\&I+L test set being blind. 
All mentioned papers do not train their systems on the development set, but use it only for model selection. Also note that we split the results on CoNLL 14 test set into two groups: those who do not use the W\&I+L data for training, and those who do. This is to allow a fair comparison, given that the W\&I+L data were not available before the BEA 2019 Shared Task on GEC.

The best performing systems are utilizing ensembles. Table~\ref{table:english_results} shows an evident performance boost (3.27-6.01 points) when combining multiple models into an ensemble. The best performing system on English is an ensemble system of \newcite{grundkiewicz2019neural}.

The aim of this paper is to concentrate on low-resource languages rather than on English. Therefore, we report results of our single model. Despite that our best system reaches 69.0 $F_{0.5}$ score, which is comparable to the performance of best systems that employ ensembles. Although \newcite{grundkiewicz2019neural} do not report their single system score, we can hypothesise that given development set scores, our system is on par with theirs or even performs slightly better.

Note that there is a significant difference between results reported on W\&I+L dev and W\&I+L test sets. This is caused by the fact that each sentence in the W\&I+L test set was annotated by 10 annotators, while there is only a single annotator for each sentence in the development set.

\subsection{German}

\newcite{boyd2018using} developed a GEC system for German based on multilayer convolutional encoder-decoder neural network~\cite{chollampatt2018mlconv}. To account for the lack of annotated data, she generated additional training data from Wikipedia edits, which she filtered to match the distribution of the original error types. As Table~\ref{table:german_results} shows, her best system reaches 45.22 $F_{0.5}$ score on Falko-Merlin test set. All our three systems outperform it.

Compared to \newcite{boyd2018using}, our system trained solely on synthetic data has lower recall, but substantially higher precision. The main reason behind the lower recall is the unsupervised approach to synthetic data generation. Both our finetuned models outperform \newcite{boyd2018using} system by a large margin.

\subsection{Czech}

\begin{table*}[t]
  \begin{center}
    \begin{tabular}{l||c||c||c}
    System & P & R  & $F_{0.5}$ \\\hline
    \newcite{rozovskaya2019grammar} & 38.0~~ & 7.5 & 21.0~~ \\\hline\hline
      Our work -- synthetic pretrain & 47.76 & 26.08 & 40.96 \\\hline 
      Our work -- finetuned base single GPU & 59.13 & 26.05 & 47.15 \\\hline
      Our work -- finetuned & 63.26 & 27.50 & 50.20 \\\hline
    \end{tabular}
  \end{center}
  \caption{Results on on RULEC-GEC Test Set (Russian).}
  \label{table:russian_results}
\end{table*}

We compare our system with \newcite{richter2012korektor}, who developed a statistical spelling corrector for Czech. Although their system can only make local changes (e.g., cannot insert a new word or swap two nearby words), it achieves surprisingly solid results. Nevertheless, all our three system perform better in both precision, recall and $F_{0.5}$ score. Possibly due to already quite high precision of the pretrained model, the finetuning stage improves mainly model recall.

We also evaluate performance of our best system on three subsets of the AKCES-GEC test set: Foreigners--Slavic, Foreigners--Other and Romani. As the name suggests, the first of them is a part of AKCES-GEC collected from essays of non-Czech Slavic people, the second from essays of non-Czech non-Slavic people and finally Romani comes from essays of Romani pupils with Romani ethnolect of Czech as their first language. The best result is reached on Romani subset, while on Foreigners--Other the $F_{0.5}$ score is by more than 6 points lower. We hypothesize this effect is caused by the fact, that Czech is the primary language of Romani pupils. Furthermore, we presume that foreigners with Slavic background should learn Czech faster than non-Slavic foreigners, because of the similarity between their mother tongue and Czech. This fact is supported by Table~\ref{tab:akcesgec_sizes}, which shows that the average error rate of Romani development set is 21.0\%, Foreigners--Slavic 21.8\% and the Foreigners--Other 23.8\%.

Finally, we report recall of the best system on each error type annotated by the first annotator (ID 0) in Figure~\ref{fig:czech_error_recalls}. Generally, our system performs better on errors annotated on Tier 1 than on errors annotated on Tier 2. 
Furthermore, a natural hypothesis is that the more occurrences there are for an error type, the better the recall of the system on the particular error type. Figure~\ref{fig:czech_error_recalls} suggests that this hypothesis seems plausible on Tier~1 errors, but its validity is unclear on Tier 2.

\subsection{Russian}

As Table~\ref{table:russian_results} indicates, GEC in Russian currently seems to be the most challenging task. Although our system outperforms the system of \newcite{rozovskaya2019grammar} by more than 100\% in $F_{0.5}$ score, its performance is still quite poor when compared to all previously described languages. Because the result of our system trained solely on synthetic data is comparable with the similar system for English, we hypothesise that the main reason behind these poor results is the small amount of annotated training data -- while Czech has 42\,210 and German 19\,237 training sentence pairs, there are only 4\,980 sentences in the Russian training set. To validate this hypothesis, we extended the original training set by 2\,000 sentences from the development set, resulting in an increase of 3 percent points in $F_{0.5}$ score.

\section{Conclusion}

We presented a new dataset for grammatical error correction in Czech. It contains almost twice as much sentences as existing German dataset and more than three times as RULEC-GEC for Russian. The dataset is published in M2 format containing both separated edits and their error types.

Furthermore, we performed experiments on three low-resource languages: German, Russian and Czech. For each language, we pretrained Transformer model on synthetic data and finetuned it with a mixture of synthetic and authentic data. On all three languages, the performance of our system is substantially higher than results of the existing reported systems. Moreover, all our models supersede reported systems even if only pretrained on unsupervised synthetic data.

The performance of our system could be even higher if we trained multiple models and combined them into an ensemble. We plan to do that in future work. We also plan to extend our synthetic corpora with data modified by supervisedly extracted rules. We hope that this could help especially in case of Russian, which has the lowest amount of training data.

\section*{Acknowledgments}

The work described herein has been supported by OP VVV VI LINDAT/CLARIN project (CZ.02.1.01/0.0/0.0/16\_013/0001781) and it has been supported and has been using language resources developed by the LINDAT/CLARIN project (LM2015071) of the Ministry of Education, Youth and Sports of the Czech Republic. This research was also partially supported by SVV project number 260~453, GAUK 578218 of the Charles University and FP7-ICT-2010-6-257528 (MŠMT 7E11042).

%

\bibliography{emnlp2018}
\bibliographystyle{acl_natbib}

\end{document}